\documentclass{article}

\usepackage{PRIMEarxiv}

\usepackage[utf8]{inputenc}
\usepackage[T1]{fontenc}
\usepackage{cite}
\usepackage{hyperref}
\usepackage{url}
\usepackage{booktabs}
\usepackage{amsfonts}
\usepackage{nicefrac}
\usepackage{microtype}
\usepackage{fancyhdr}
\usepackage{graphicx}
\usepackage{array}
\usepackage{float}
\graphicspath{{media/}}
\title{Estimating the Impact of COVID-19 on Travel Demand in Houston Area Using Deep Learning and Satellite Imagery}

\author{
  Alekhya Pachika$^{1}$, Lu Gao, Ph.D.$^{2}$, Lingguang Song, Ph.D.$^{3}$ \\
  Civil and Environmental Engineering, University of Houston \\
  \texttt{apachika@uh.edu, lgao5@central.uh.edu, lsong5@central.uh.edu} \\
  \And
  Pan Lu, Ph.D.$^{4}$ \\
  $^{4}$Department of Transportation, Logistics and Finance, North Dakota State University \\
  \texttt{pan.lu@ndsu.edu} \\
  \And
  Xingju Wang, Ph.D.$^{5}$ \\
  $^{5}$School of Traffic and Transportation, Shijiazhuang Tiedao University \\
  \texttt{wangxingju@stdu.edu.cn} \\
}

\begin{document}
\maketitle

\begin{abstract}
Considering recent advances in remote sensing satellite systems and computer vision algorithms, many satellite sensing platforms and sensors have been used to monitor the condition and usage of transportation infrastructure systems. The level of detail that can be detected increases significantly with the increase of ground sample distance (GSD), which is around 15 cm--30 cm for high-resolution satellite images. In this study, we analyzed data acquired from high-resolution satellite imagery to provide insights, predictive signals, and trends for travel demand estimation. More specifically, we estimate the impact of COVID-19 in the metropolitan area of Houston using satellite imagery from Google Earth Engine datasets. We developed a car-counting model through Detectron2 and Faster R-CNN to monitor the presence of cars within different locations (i.e., university, shopping mall, community plaza, restaurant, supermarket) before and during COVID-19. The results show that the number of cars detected at these selected locations reduced on average 30\% in 2020 compared with the previous year 2019. The results also show that satellite imagery provides rich information for travel demand and economic activity estimation. Together with advanced computer vision and deep learning algorithms, it can generate reliable and accurate information for transportation agency decision makers.
\end{abstract}


\section{Introduction}

Satellite image analysis uses images taken from an artificial satellite and analyzes them for various purposes such as meteorology, landscape, regional planning, agricultural studies, geology (geomorphology), forestry, urban studies, geography, environmental research, cartography, aerology, climatology, oceanography, military, intelligence, and warfare~\cite{jean2016view, henderson2015night}. The advancement of satellites and image-analysis methods has become more sophisticated in the past decade and has led to many applications~\cite{gorelick2017earthengine, nussbaum2023cityscale, wang2020wuhantraffic}. Recent years have seen an upsurge in interest in high-resolution synthetic aperture radar (SAR) data and innovative data processing methods. Thanks to these developments, it has now become possible to track issues in relatively small targets and their extent. For this reason, satellite imagery analysis has been extensively used for applications in civil engineering, including monitoring the condition of infrastructure facilities, detecting damage, evaluating impacts of disasters, and monitoring the usage of infrastructure facilities~\cite{lebaku2024deep, wang2024comprehensive}.

The research community has discussed the possibility of applying satellite imagery for pavement management in the past decade. For example, Haider et al.~\cite{haider2010} suggested that a satellite-based system of pavement monitoring could improve highway maintenance and reduce the number of vehicle-based inspections. Faghri and Ozden~\cite{faghri2015} reviewed and summarized the fields that use satellite imagery. This study examines how effective satellite imagery is in pavement management, analyzing historical data and identifying deformations and deformation velocity for highways, railways, and pavement roughness. Li et al.~\cite{li2017} studied the financial aspects of the technology and its components based on a cost-benefit analysis of ongoing pavement monitoring activities. More recently, due to the fast development of deep learning-based image analysis tools, researchers have investigated the effectiveness of satellite imagery in monitoring pavement surface conditions~\cite{cheng2020benchmark, cheng2016survey, kellenberger2020review, shi2022survey, li2019scene, zhao2019generic, zou2019survey, long2021millionaid, liu2021method}. Brewer et al.~\cite{brewer2021} summarized the use of remotely sensed images to determine road quality with convolutional neural networks. As inputs, they used high-resolution satellite imagery to assemble information about road quality and achieved 80\% accuracy. Bashar and Torres-Machi~\cite{bashar2022} summarized that satellite imagery can be used as a cost-effective and rapid means for evaluating the condition of roads. An assessment to find pavement distresses was conducted using spectral and texture information derived from 30-cm panchromatic and 1.2-m optical imagery. Based on the spectral analysis of the multispectral images, it was found that the smoother the roads, the brighter the pavement across the whole spectrum. The spectral index approach may be useful for identifying individual distresses; however, the opposing behavior of pixel brightness values limits its application to analyzing the condition of a pavement when multiple distresses are present.

Satellite imagery has also been applied to monitor building damage. For example, Wang et al.~\cite{wang2009} used SAR and optical images from the Wenchuan earthquake to determine a damaged building's characteristics. Pan and Tang~\cite{pan2010} studied the interrelationships between grade of building damage and variations in backscatter intensity in the case of the Wenchuan earthquake. Three categories of damage were identified: severely damaged, moderately damaged, and somewhat damaged. Dell'Acqua and Polli~\cite{dellacqua2011} analyzed the differences in texture statistics based solely on post-event COSMO/SkyMed data to assess damage based only on radar reflectivity patterns affected by damage based on block-averaged measurements. Cossu et al.~\cite{cossu2012} summarized how SAR images of various resolutions can be used to examine damage and textural characteristics. In this study, the authors used a completely diverse training sample from different spatial resolutions in satellite images to classify building damage more accurately. Satellite imagery analysis through deep learning models is increasingly being used for monitoring building damage in recent studies. For example, Ji et al.~\cite{ji2018} used a convolutional neural network to identify collapsed buildings in the Haiti earthquake with an overall accuracy of 78.6\%. Nex et al.~\cite{nex2019} stated that deep learning techniques have improved traditional image analysis approaches, allowing them to accurately identify visible structural damage in buildings. Their work investigated the performance of a CNN for detecting visible structural damage. It is evident from the experiments presented that many factors impact the quality of the results and that it is nearly impossible to predict the exact behavior of the network with a dataset in advance, especially when the dataset includes different geographical regions and different building types compared to the ones used as training samples.

Since satellite imagery has developed dramatically in recent years, studies have concentrated more on how satellite imagery can be used for monitoring the usage of transportation infrastructure systems~\cite{jin2006vehicle, lefebvre2007vehicle, wu2013parkinglot, mundhenk2015lowres, dunford2017multimodal, moranduzzo2017segment, aksoy2018comprehensive, du2019comparison, vanetten2020vehiclevessel, solano2020vehsat, cai2021targetguided, li2022densitymap, yao2019parkinglot, miyanaga2019hierarchical, alsolami2023realtime, ha2023improved}. For example, Hoppe et al.~\cite{hoppe2016} summarized that monitoring long-term transport infrastructure with InSAR technology is appealing because of the wide availability of radar satellites and the rapid development of digitized signal processing techniques. Chen et al.~\cite{chen2021, gao2022evaluating, pachika2023estimating} used multitemporal planet satellite images and developed a vehicle detection method to determine how mobility has changed in numerous cities around the world because of the COVID-19 pandemic.

\section{Estimating COVID-19 Impact on Travel Demand}

There is a growing interest in the analysis of satellite imagery in the domain of monitoring usage of infrastructure facilities. When governments need to monitor economic activities in cities, they can act fast and efficiently through satellite imagery analysis. The recurrence of hurricanes and the COVID-19 pandemic has caused crises with impacts on local economic sectors. The COVID-19 pandemic has also created substantial disruptions across multiple dimensions of transportation systems and infrastructure decision-making. In this case study, we propose an automatic approach for analyzing the impact on economic activities via satellite images. In this application, we apply the Detectron2 object detection pipeline~\cite{wu2019} together with the Faster R-CNN family of detectors~\cite{ren2015faster} to the task of car detection in satellite imagery. The car detection model was trained on a very large and diverse satellite image dataset with around 30 cm resolution. The model was able to achieve 90\% accuracy on average. This journal manuscript further extends our earlier conference investigation of COVID-19-related travel demand changes in Houston using satellite imagery and deep learning.

\subsection{Car Detection Model}

Computer vision techniques such as object detection allow us to identify and locate items in images or videos. The precise location of objects in a scene can be determined through this kind of identification and localization, while the objects can be counted and accurately categorized. Similar deep learning approaches have also been applied to traffic-state evaluation and prediction in transportation systems.

\begin{figure}[H]
  \centering
  \includegraphics[width=0.96\textwidth]{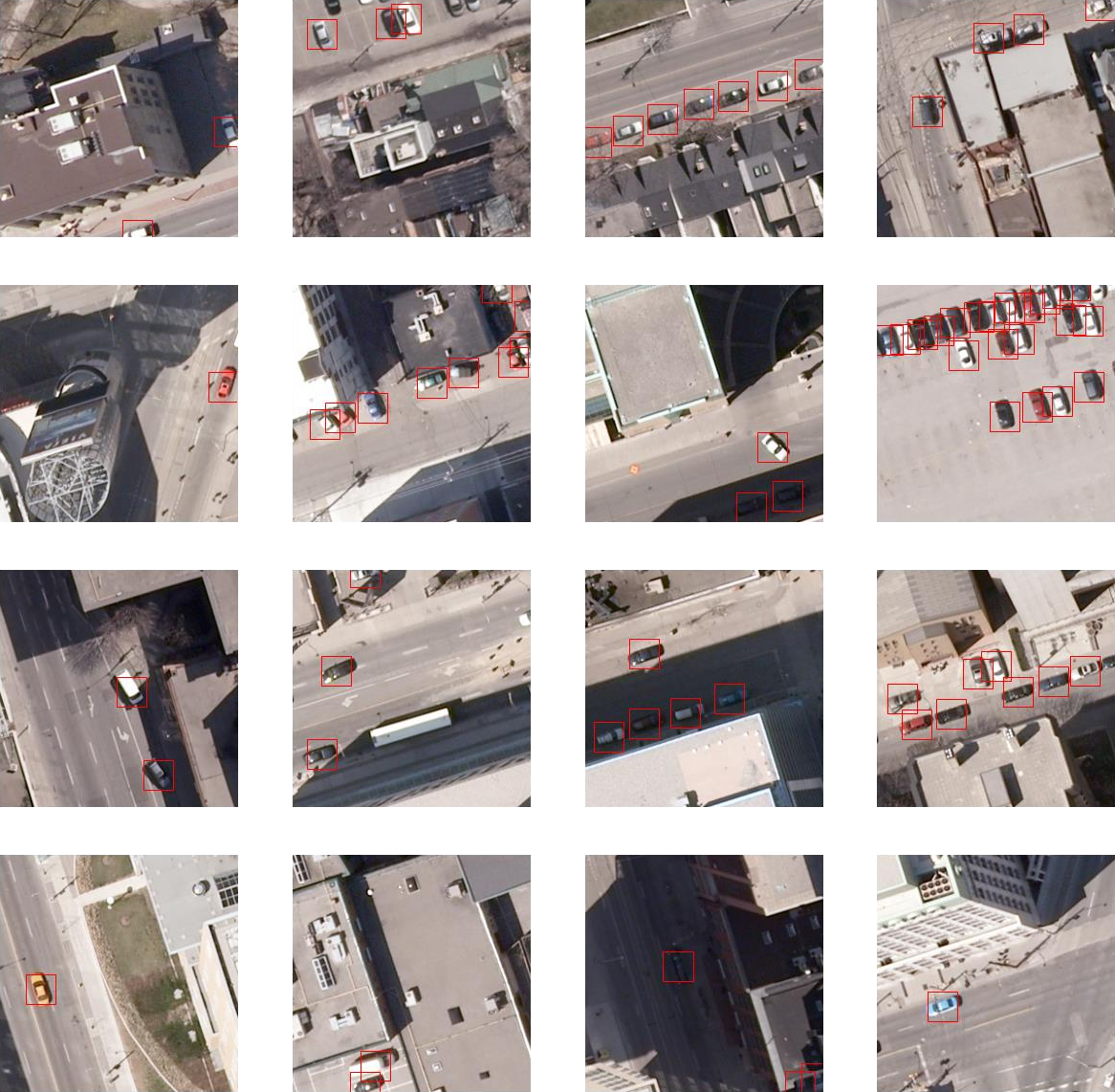}
  \caption{Sixteen COWC images and bounding boxes.}
  \label{fig:cowc}
\end{figure}

In this case study, we trained our vehicle counting model using the Cars Overhead with Context (COWC) dataset, which consists of annotated cars from satellite imagery~\cite{mundhenk2016}. The data consists of around 33,000 unique cars from six different image locations: Toronto, Canada; Selwyn, New Zealand; Potsdam and Vaihingen, Germany; and Columbus and Utah, United States. The Columbus and Vaihingen datasets are in grayscale, which are not used for the training in this case study. The other datasets are 3-band RGB images. The COWC imagery has a resolution of around 15 cm ground sample distance (GSD). We split the dataset into 70\% training and 30\% testing. The bounding boxes of 16 sample COWC images are displayed in Figure~\ref{fig:cowc}. We trained the Detectron2 model for 9,000 epochs, which takes about 1 hour on a Google Colab Pro account.

\subsection{Dataset: High-Resolution Dataset over Houston}

In this section, we collected multiple images over Houston using Google Earth Pro and the National Oceanic and Atmospheric Administration (NOAA) website. We downloaded images from eight different parking spaces in Houston. In total, we collected 127 images over the past few years.

\begin{table}[H]
  \caption{Data collected from 8 places in Houston.}
  \label{tab:houston-data}
  \centering
  \begin{tabular}{>{\raggedright\arraybackslash}p{0.08\textwidth} >{\raggedright\arraybackslash}p{0.70\textwidth} >{\centering\arraybackslash}p{0.12\textwidth}}
    \toprule
    No. & Place & Area (km$^2$) \\
    \midrule
    1 & Evelyn Rubenstein Jewish Community Center (JCC) of Houston & 0.04 \\
    2 & University of Houston (UH) Parking Lot 16B, 16C, and 16F & 0.04 \\
    3 & Southwest corner of the intersection of Chimney Rock Rd and S Braeswood Blvd & 0.02 \\
    4 & Braeswood Square & 0.06 \\
    5 & South of the intersection of Hillcroft Ave and S Braeswood Blvd & 0.07 \\
    6 & Meyerland Plaza & 0.21 \\
    7 & Houston Chinatown & 0.40 \\
    8 & Katy Mills & 0.51 \\
    \bottomrule
  \end{tabular}
\end{table}

The selection of these locations in Houston allows us to analyze different economic sectors. For example, locations 1, 3, 4, and 5 can be used to assess the impacts of flooding and COVID-19 on community commercial centers. Location 2 can be used to analyze the effect of COVID-19 on university students' attendance. Locations 6 and 8 can be used to measure economic activity at major commercial malls. Location 7 can be used to measure the impacts of disasters on local restaurants.

\begin{figure}[H]
  \centering
  \includegraphics[width=\textwidth]{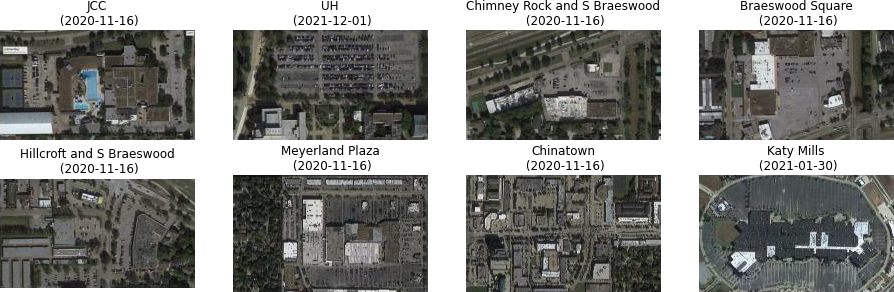}
  \caption{Locations selected for this case study.}
  \label{fig:locations}
\end{figure}

The following figures show the comparison between the periods before and during COVID-19. We can visually observe a reduction in the total number of cars before and during the pandemic.

\begin{figure}[H]
  \centering
  \includegraphics[width=\textwidth]{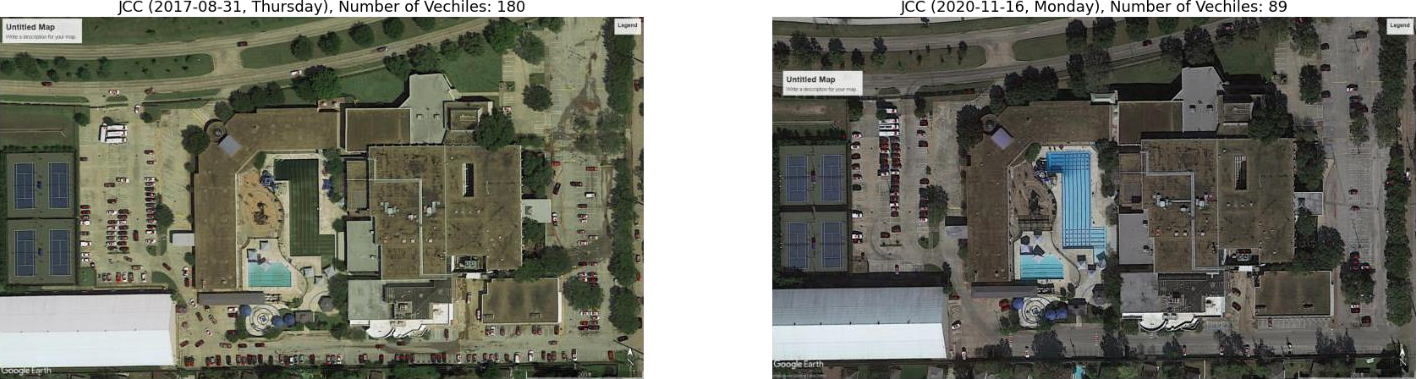}
  \caption{Parking located at JCC.}
\end{figure}

\begin{figure}[H]
  \centering
  \includegraphics[width=\textwidth]{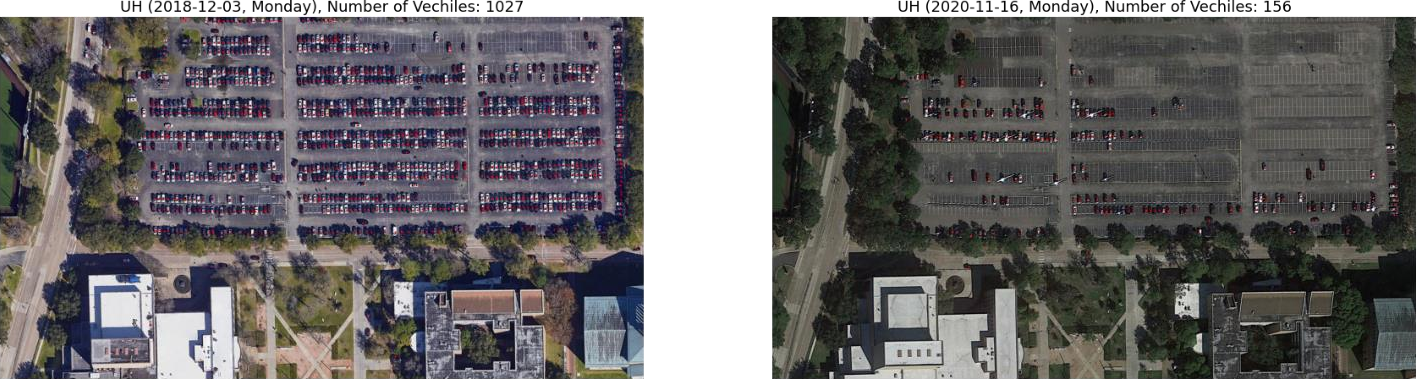}
  \caption{Parking located at UH.}
\end{figure}

\begin{figure}[H]
  \centering
  \includegraphics[width=\textwidth]{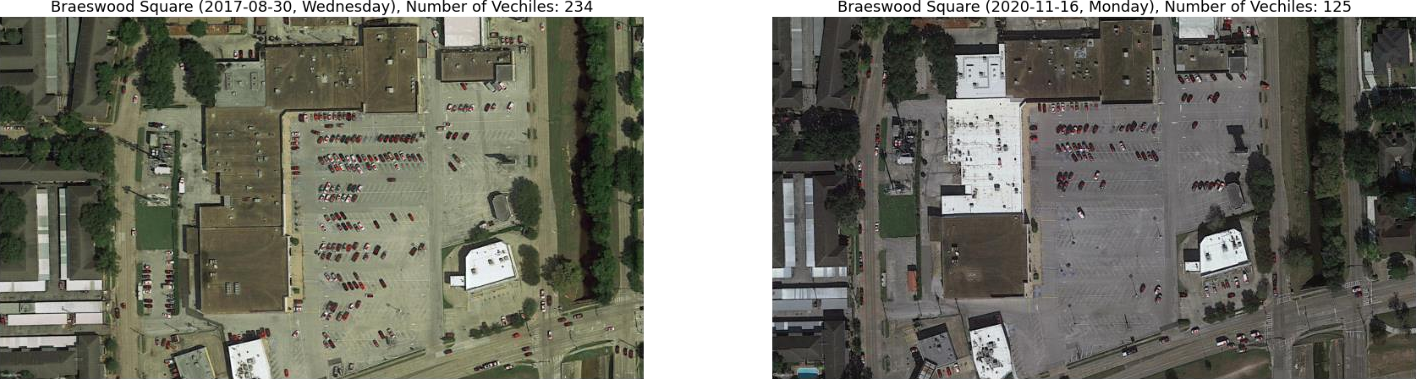}
  \caption{Parking located at Braeswood Square.}
\end{figure}

\begin{figure}[H]
  \centering
  \includegraphics[width=\textwidth]{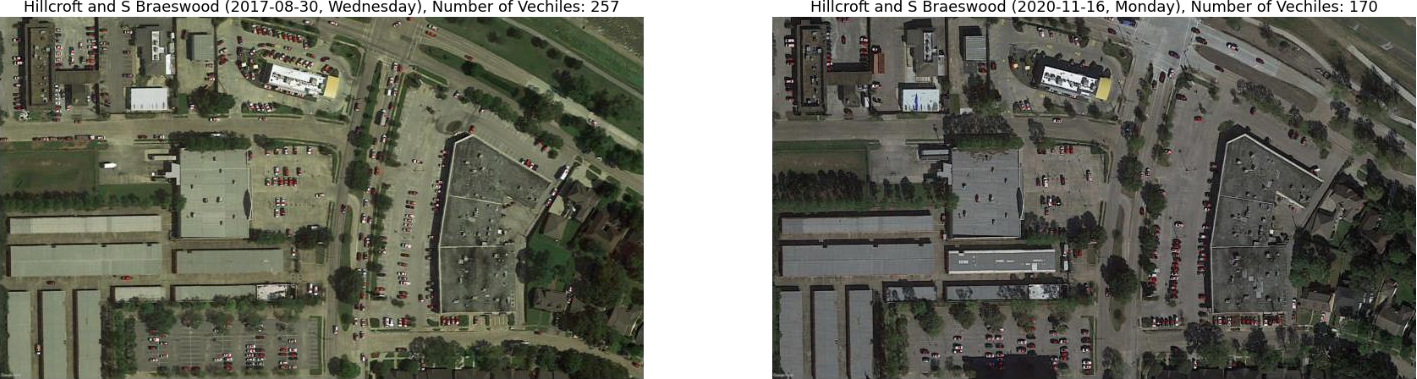}
  \caption{Parking located at Hillcroft and S Braeswood.}
\end{figure}

\begin{figure}[H]
  \centering
  \includegraphics[width=\textwidth]{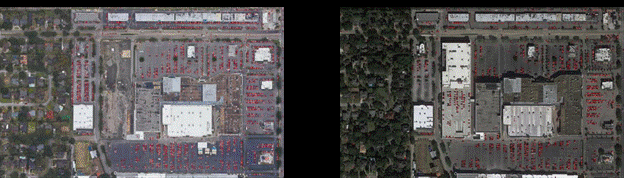}
  \caption{Parking located at Meyerland Plaza.}
\end{figure}

\begin{figure}[H]
  \centering
  \includegraphics[width=\textwidth]{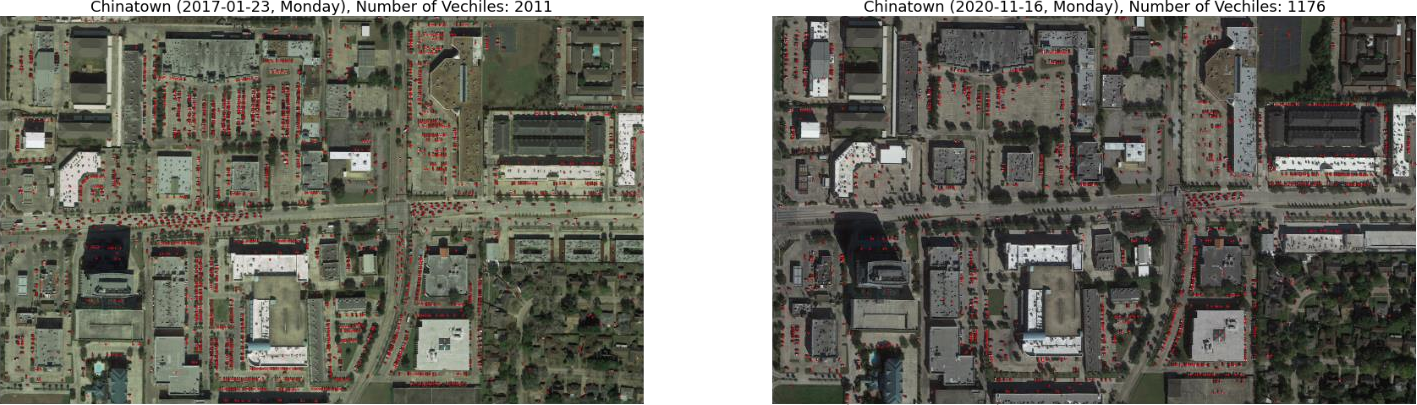}
  \caption{Parking located at Chinatown.}
\end{figure}

\begin{figure}[H]
  \centering
  \includegraphics[width=\textwidth]{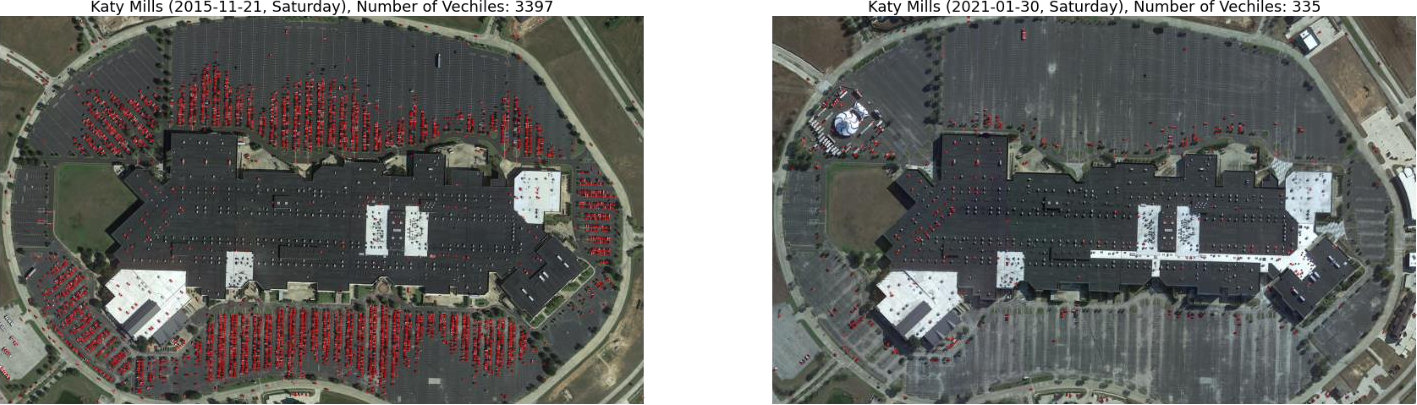}
  \caption{Parking located at Katy Mills.}
\end{figure}

By counting the number of vehicles in a certain region during the past decade, we were able to measure changes in economic activity. We cropped each collected satellite image into tiles of size 256$\times$256 and passed them through the Detectron2 pipeline. Overlapped bounding boxes were discarded when combining tiles together to remove repeated vehicles in consecutive tiles. The results show that the number of cars parked at these selected locations reduced on average 30\% in 2020 compared with the previous year 2019. These findings also suggest that image-derived transportation indicators can complement broader infrastructure and socio-economic assessment frameworks.

\section{Conclusion}

This research reviewed previous studies on satellite image analysis and its application in infrastructure management, including monitoring pavement conditions, disaster management, and damage assessment. As a result, the following conclusions have been summarized:

\begin{itemize}
\item Satellite image analysis provides cost-effective approaches for continuously monitoring infrastructure assets that cover large areas. As a complement to traditional methods and practices, satellite-based infrastructure monitoring is useful.
\item The studies have demonstrated that convolutional neural network (CNN) related deep learning models are promising in estimating road quality from remotely sensed imagery. Further research is needed to better understand their applications in pavement quality estimation. With additional research, this method could be valuable in estimating pavement quality.
\item The application of satellite imagery methods to sinkhole detection, slope stability monitoring, and building damage monitoring has been proven effective. The ability of satellite imagery methods to measure surface displacements at the millimeter scale over a large landmass creates the potential for network-level implementation.
\item The increased availability of radar satellites, combined with the rapid progress in digital signal processing, is useful for long-term performance monitoring of infrastructure systems.
\item The images used in the case study were taken from Google Earth Pro, which only provides a few images each year. This may add uncertainty in estimating the number of vehicles. For future studies, more images from commercial satellite vendors are needed to overcome this limitation. Another approach to reduce the uncertainty is to estimate yearly average vehicle counts.
\item The economic impact can be better estimated if other data sources such as surveys, interviews, employment, and prices are integrated together with satellite image analysis.
\end{itemize}

\bibliographystyle{unsrt}
\bibliography{references_all}

\end{document}